\documentclass{modified-kluwer}

\newdisplay{guess}{Conjecture}

\usepackage{graphicx}

\begin{document}
\begin{article}
\begin{opening}

\title{Corpus-based Learning of Analogies \\ and Semantic Relations}
\author{Peter D. \surname{Turney} \email{peter.turney@nrc-cnrc.gc.ca}}
\institute{Institute for Information Technology \\
National Research Council Canada \\
M-50 Montreal Road \\
Ottawa, Ontario, Canada, K1A 0R6 \\
phone: (613) 993-8564 \\
fax: (613) 952-7151}
\author{Michael L. \surname{Littman} \email{mlittman@cs.rutgers.edu}}
\institute{Department of Computer Science \\
Rutgers University \\
Piscataway, NJ 08854-8019, USA}
\runningauthor{Turney and Littman}
\runningtitle{Corpus-based Learning of Analogies and Semantic Relations}
\date{\today}

\begin{abstract}
We present an algorithm for learning from unlabeled text, based on the
Vector Space Model (VSM) of information retrieval, that can solve verbal
analogy questions of the kind found in the SAT college entrance exam.
A verbal analogy has the form {\em A:B::C:D\/}, meaning
``\(A\) is to \(B\) as \(C\) is to \(D\)'';
for example, mason:stone::carpenter:wood. SAT analogy questions provide
a word pair, {\em A:B\/}, and the problem is to select the most analogous word
pair, {\em C:D\/}, from a set of five choices. The VSM algorithm correctly
answers 47\% of a collection of 374 college-level analogy questions
(random guessing would yield 20\% correct; the average college-bound senior
high school student answers about 57\% correctly). We motivate this research
by applying it to a difficult problem in natural language processing,
determining semantic relations in noun-modifier pairs. The problem
is to classify a noun-modifier pair, such as ``laser printer'', according
to the semantic relation between the noun (printer) and the modifier
(laser). We use a supervised nearest-neighbour algorithm that assigns
a class to a given noun-modifier pair by finding the most analogous
noun-modifier pair in the training data. With 30 classes of semantic
relations, on a collection of 600 labeled noun-modifier pairs, the
learning algorithm attains an \(F\) value of 26.5\% (random guessing: 3.3\%).
With 5 classes of semantic relations, the \(F\) value is 43.2\% (random: 20\%).
The performance is state-of-the-art for both verbal analogies and noun-modifier relations.
\end{abstract}


\end{opening}

\section{Introduction}\label{sec:intro}

Computational approaches to analogy-making typically involve hand-coded knowledge
bases \cite{Fre}. In this paper, we take a different approach, based on the
idea that analogical reasoning can be approximated to some extent by
a cosine measure of vector similarity, where the vectors are derived from
statistical analysis of a large corpus of text. We demonstrate this approach with
two real-world problems, answering multiple-choice verbal analogy questions
and classifying noun-modifier semantic relations. This work is only a first step,
and analogical reasoning is still very far from being a solved problem,
but we believe that our results are encouraging. A vector-based approach
to analogies and semantic relations may be able to overcome some of the limitations
(such as the knowledge-engineering bottleneck) that have impeded progress with the 
knowledge-based approach.

A verbal analogy has the form {\em A:B::C:D\/}, meaning
``\(A\) is to \(B\) as \(C\) is to \(D\)''; for example,
``mason is to stone as carpenter is to wood''. (A mason is an artisan who works with
stone; a carpenter is an artisan who works with wood.) Analogies of this kind
are sometimes called {\em proportional analogies}, and they have 
been studied at least since 350~BC \cite{Ari}.
In spite of their long history, they are still not well understood; their
subjective character resists quantitative analysis. In our research, we
have used multiple-choice questions, developed for educational testing, as a tool
for objective analysis of verbal analogies.

The SAT college entrance exam
contains multiple-choice verbal analogy questions, in which there is a word pair,
{\em A:B\/}, and five choices.\endnote{The College Board has announced that analogies
will be eliminated from the SAT in 2005
(http://www.collegeboard.com/about/newsat/newsat.html),
as part of a shift in the exam to reflect changes in the curriculum. The SAT was
introduced as the Scholastic Aptitude Test in 1926, its name was
changed to Scholastic Assessment Test in 1993, then changed to simply
SAT in 1997.} The task is to select the most analogous
word pair, {\em C:D\/}, from the set of five word pairs. Table~\ref{sample} gives an
example. In the terminology of educational testing, the
first pair, {\em A:B\/}, is called the {\em stem} of the analogy.

\begin{table}[b]
\begin{tabular}{lrl}
\hline
Stem:      &       & mason:stone        \\
\hline
Choices:   & (a)   & teacher:chalk      \\
           & (b)   & carpenter:wood     \\
           & (c)   & soldier:gun        \\
           & (d)   & photograph:camera  \\
           & (e)   & book:word          \\
\hline
Solution:  & (b)   & carpenter:wood     \\
\hline
\end{tabular}
\vspace{-5pt}
\caption[]{A sample SAT question.}\label{sample}
\end{table}

For multiple-choice analogy questions, the best choice is the word pair with the semantic
relation that is most similar to the relation of the stem pair. Although there has been much
research on measuring the similarity of {\em individual concepts}
\cite{Les, ChuHan, Dun, Sma, Res, LanDum, Tur, PanLin},
there has been relatively little work on measuring
the similarity of {\em semantic relationships} between concepts
\cite{Van, RosHea, Ros, NasSzp}.

Our approach to verbal analogies is inspired by the Vector Space Model (VSM) of
information retrieval \cite{SalMcG, Sal}.
We use a vector of numbers to
represent the semantic relation between a pair of words. The similarity between two
word pairs, {\em A:B\/} and {\em C:D\/}, is measured by the cosine of the angle
between the vector that represents {\em A:B\/} and the vector that represents
{\em C:D\/}. 

As we discuss in Section~\ref{subsec:vsm}, the VSM
was originally developed for use in information retrieval. Given a query, a set of documents can
be ranked by the cosines of the angles between the query vector and each document
vector. The VSM is the basis for most modern search engines
\cite{BaeRib}.

Section~\ref{sec:relwork} also covers related work on analogy, metaphor, and classifying
semantic relations. Most of the related work has used manually constructed lexicons and
knowledge bases. Our approach uses learning from unlabeled text, with a very large
corpus of web pages (about one hundred billion words); we do not use a lexicon or
knowledge base.

We present the details of our learning algorithm in Section~\ref{sec:solving}, including an experimental
evaluation of the algorithm on 374 college-level SAT-style verbal analogy questions. The
algorithm correctly answers 47\% of the questions. Since there are five choices per
analogy question, random guessing would be expected to result in 20\% correctly
answered. We also discuss how the algorithm might be extended from {\em recognizing}
analogies to {\em generating} analogies.

To motivate research on verbal analogies, we give an example of a practical application,
the task of classifying the semantic relations of noun-modifier
pairs. Given a noun-modifier pair such as ``laser printer'', the problem is to classify the
semantic relation between the noun (printer) and the modifier (laser). In 
Section~\ref{subsec:nounmodapps}, we argue that an algorithm
for classification of noun-modifier relations would be useful in machine translation,
information extraction, and word sense disambiguation.

An algorithm for solving SAT-style verbal analogies can be applied to classification of
noun-modifier semantic relations, as we demonstrate in Section~\ref{sec:nounmods}. Given an unclassified 
noun-modifier pair, we can search through a set of labeled training data for the most 
analogous noun-modifier pair. The idea is that the class of the nearest neighbour in the 
training data will predict the class of the given noun-modifier pair.
We apply a supervised nearest-neighbour learning algorithm, where the measure of
distance (similarity) is the cosine of the vector angles. 

The data set for the experiments in Section~\ref{sec:nounmods}
consists of 600 labeled noun-modifier pairs, from \citeauthor{NasSzp} (\citeyear{NasSzp}).
The learning algorithm attains an \(F\) value of 26.5\% when given 30 different classes of
semantic relations. Random guessing would be expected to result in an \(F\) value of 3.3\%.
We also consider a simpler form of the data, in which the 30 classes have been collapsed
to 5 classes. The algorithm achieves an \(F\) value of 43.2\% with the 5-class version of the
data, where random guessing would be expected to yield 20\%.

Limitations and future work are covered in Section~\ref{sec:future}. The conclusion follows in
Section~\ref{sec:conclusion}.

\section{Related Work}\label{sec:relwork}

In this section, we consider related work on metaphorical and analogical reasoning
(Section~\ref{subsec:metana}), applications of the Vector Space Model (Section~\ref{subsec:vsm}), 
and research on classifying noun-modifier pairs according to their semantic relations 
(Section~\ref{subsec:semrels}). We also discuss related work on web mining for natural
language processing applications (Section~\ref{subsec:webmining}).

\subsection{Metaphor and Analogy}\label{subsec:metana}

\citeauthor{Turetal} (\citeyear{Turetal})
presented an ensemble approach to solving verbal analogies.
Thirteen independent modules were combined using three different merging rules. One of
the thirteen modules was the VSM module, exactly as presented here in Section~\ref{subsec:vsmapproach}.
However, the focus of \citeauthor{Turetal} (\citeyear{Turetal})
was on the merging rules; the individual
modules were only briefly outlined. Therefore it is worthwhile to focus here on the VSM
module alone, especially since it is the most accurate of the thirteen modules. 
Table~\ref{ensembles} shows the impact of the VSM module on the accuracy of the ensemble.
These figures suggest that the VSM module
made the largest contribution to the accuracy of the ensemble. The present paper goes
beyond \citeauthor{Turetal} (\citeyear{Turetal})
by giving a more detailed description of the VSM module,
by showing how to adjust the balance of precision and recall, and by examining the
application of the VSM to the classification of noun-modifier relations.

\begin{table}[b]
\begin{tabular}{lrr}
\hline
                           & With VSM           & Without VSM           \\
\hline
Ensemble accuracy          & 45.0\%             & 37.0\%                \\
Best individual module     & VSM                & Wordsmyth Similarity  \\
Best individual accuracy   & 38.2\%             & 29.4\%                \\
\hline
\end{tabular}
\vspace{-5pt}
\caption[]{Impact of the VSM module on ensemble accuracy.}\label{ensembles}
\end{table}

\citeauthor{Fre} (\citeyear{Fre})
surveyed the literature on computational modeling of analogy-making.
The earliest work was a system called Argus, which could solve a few simple verbal
analogy problems \cite{Rei}. Argus used a small hand-built semantic network and
could only solve the limited set of analogy questions that its programmer had anticipated.
All of the systems surveyed by French used hand-coded knowledge-bases; none of them
can learn from data, such as a corpus of text. 

\citeauthor{Fre} (\citeyear{Fre}) cited Structure Mapping Theory (SMT)
\cite{Get} and its implementation in the Structure Mapping Engine (SME)
\cite{Faletal} as the most influential work on modeling of analogy-making.
SME takes representations of a source domain and a target domain,
and produces an analogical mapping between the source and target.
The domains are given structured propositional representations, using
predicate logic. These descriptions include {\em attributes} (unary 
predicates indicating features), {\em relations} (expressing connnections between 
entities), and {\em higher-order relations} (expressing connections between relations).
The analogical mapping connects source domain relations
to target domain relations. Originally, only identical relations were mapped, but
later versions of SME allowed similar, non-identical relations to match
\cite{Fal}. 

With proportional analogies of the form {\em A:B::C:D\/}, the target
and source domains are reduced to a minimum. Each domain consists
of two features (explicitly given) and one relation (implicit), $R_1(A,B)$ and
$R_2(C,D)$. The focus in our work is on the
similarity measure that is used to compare the relations, rather than the
analogical mapping process. We believe it is a good research strategy to focus on
one aspect of the problem at a time, mapping between complex predicate 
logic structures \cite{Faletal}
or measuring similarity of relations (as we do here), but eventually 
researchers will need to address both problems together.
Real-world analogies involve domains
with complex internal structures and complicated relational similarities.

\citeauthor{Dol} (\citeyear{Dol})
described a system for extracting semantic information from machine-readable
dictionaries. Parsing and semantic analysis were used to convert the Longman
Dictionary of Contemporary English (LDOCE) into a large Lexical Knowledge Base
(LKB). The semantic analysis recognized twenty-five different classes of semantic
relations, such as {\em hypernym (is\_a), part\_of, typical\_object, means\_of},
and {\em location\_of}. \citeauthor{Dol} (\citeyear{Dol})
outlined an algorithm for identifying ``conventional'' metaphors in the LKB.
A {\em conventional} metaphor is a metaphor that is familiar to a native speaker and has
become part of the standard meaning of the words involved \cite{LakJoh}.
For example, English speakers are familiar with the metaphorical links between
(sporting) games and (verbal) arguments. Dolan's algorithm can identify this
metaphorical connection between ``game'' and ``argument'' by observing the similarity in
the LKB of the graph structure in the neighbourhood of ``game'' to the graph structure in
the neighbourhood of ``argument''. The examples of metaphors identified by the
algorithm look promising, but the performance of the algorithm has not been objectively
measured in any way (e.g., by SAT questions). Unfortunately, the LKB and the
algorithms for parsing and semantic analysis are proprietary, and are therefore not
evaluated in the present paper.

The VSM algorithm is not limited to conventional metaphors. For example,
the analogy tourniquet:bleeding::antidote:poisoning was discovered by
the VSM approach (see Section~\ref{subsubsec:generating}).

\citeauthor{Vea} (\citeyear{Vea})
presented an algorithm for automatically enhancing WordNet \cite{Fel} to facilitate
analogical reasoning. The algorithm adds new links to the WordNet graph structure
by analyzing the glosses (definitions). The algorithm was designed with a 
focus on analogies of the form adjective:noun::adjective:noun, such as:

\begin{itemize}
\item Christian:church::Muslim:mosque
\item Greek:Zeus::Roman:Jove
\item Greek:alpha::Hebrew:aleph.
\end{itemize}

\noindent \citeauthor{Vea} (\citeyear{Vea}) reported a recall of 61\% and a precision of 93.5\%
for the task of creating analogical mappings between the gods of five different
cultures (Greek, Roman, Hindu, Norse, and Celtic). It would be interesting to see whether
this approach can be extended to handle SAT questions, which are not limited
to adjective:noun pairs.\endnote{All nine possible combinations of noun, verb, and
adjective can be found in the word pairs in our 374 SAT questions. In an SAT analogy
{\em A:B::C:D\/}, $A$ and $C$ have the same part of speech and $B$ and $D$ have
the same part of speech.}

\citeauthor{Maretal} (\citeyear{Maretal}) developed an unsupervised algorithm
for discovering analogies by clustering words from two different corpora. Each
cluster of words in one corpus is coupled one-to-one with a cluster in the
other corpus. With conventional clustering, the quality of the clustering of
a set of words is typically measured by considering all possible pairs
of words. The clustering is good when pairwise similarity is high for words that
are in the same cluster and low for words that are in different clusters.
With coupled clustering, the quality of the clustering is measured by considering
all pairs of words in which one member of the pair is from the first corpus and
the other member of the pair is from the second corpus. The clustering is good when 
pairwise similarity is high for words that are in the same coupled clusters
and low for words that are in different coupled clusters. For example, one
experiment used a corpus of Buddhist documents and a corpus of Christian documents.
A cluster of words such as \{Hindu, Mahayana, Zen, ...\} from the Buddhist corpus
was coupled with a cluster of words such as \{Catholic, Protestant, ...\} from the
Christian corpus. Thus the algorithm appears to have discovered an analogical
mapping between Buddhist schools and traditions and Christian schools and traditions.
\citeauthor{Dagetal} (\citeyear{Dagetal}) extend this approach from two different
corpora to any number of different corpora. This is interesting work, but it is
not directly applicable to SAT analogies, because it discovers analogies between
clusters of words, rather than individual words. 

\citeauthor{LapLas} (\citeyear{LapLas}) described a corpus-based algorithm
for logical metonymy. Metonymy and metaphor are distinct but closely related
\cite{LakJoh}. {\em Metonymy} is referring to something by mentioning an attribute
or feature of the thing. In {\em logical metonymy}, an event is referred to by mentioning
a noun that is involved in the event. For example, in the sentence
``Mary finished the cigarette'', the implicit event is smoking the cigarette,
which is metonymically referred to by explicitly mentioning only the cigarette,
omitting reference to smoking. Given a logical metonymy as input,
the algorithm of \citeauthor{LapLas} (\citeyear{LapLas}) can produce a
list of non-metonymical paraphrases of the input sentence, sorted in order
of decreasing probability. Given ``John began the cigarette'', the system
would produce ``John began smoking the cigarette'', ``John began rolling the
cigarette'', ``John began lighting the cigarette'', and so on. This work is
related to our work in the use of a corpus-based approach, but the details
of the algorithms and the tasks are quite different.

\subsection{Vector Space Model}\label{subsec:vsm}

In information retrieval, it is common to measure the similarity between a query and a
document using the cosine of the angle between their vectors
\cite{SalMcG, Sal}. Almost all modern search engines use the VSM to rank documents by
relevance for a given query.

The VSM approach has also been used to measure the semantic similarity of words
\cite{Les, Rug, PanLin}. \citeauthor{PanLin} (\citeyear{PanLin}) clustered words
according to their similarity, as measured by a VSM. Their algorithm is able to discover
the different senses of a word using unsupervised learning. They achieved impressive
results on this ambitious task.

The novelty in our work is the application of the VSM approach to measuring the
similarity of semantic relationships. The vectors characterize the semantic relationship
between a pair of words, rather than the meaning of a single word
\cite{Les} or the topic of a document \cite{SalMcG}.

\subsection{Noun-Modifier Semantic Relations}\label{subsec:semrels}

\citeauthor{NasSzp} (\citeyear{NasSzp}) used supervised learning to classify noun-modifier
relations. To evaluate their approach, they created a set of 600 noun-modifier pairs,
which they hand-labeled with 30 different classes of semantic relations. (We use this data
set in our own experiments, in Section~\ref{sec:nounmods}.) Each noun-modifier word pair was represented
by a feature vector, where the features were derived from the ontological hierarchy in a
lexicon (WordNet or Roget's Thesaurus). Standard machine learning tools (MBL, C5.0,
RIPPER, and FOIL) were used to induce a classification model from the labeled feature
vectors. \citeauthor{NasSzp} (\citeyear{NasSzp}) described their work as exploratory; the results
they presented were qualitative, rather than quantitative. Their approach seems
promising, but it is not yet ready for a full quantitative evaluation.

\citeauthor{RosHea} (\citeyear{RosHea}) used supervised learning to classify noun-modifier
relations in the medical domain, using MeSH (Medical Subject Headings) and UMLS
(Unified Medical Language System) as lexical resources for representing each noun-modifier 
relation with a feature vector. They achieved good results using a neural
network model to distinguish 13 classes of semantic relations. In an extension of this
work, \citeauthor{Ros} (\citeyear{Ros}) used hand-crafted rules and features derived from MeSH to
classify noun-modifier pairs that were extracted from biomedical journal articles. Our
work differs from \citeauthor{RosHea} (\citeyear{RosHea}) and
\citeauthor{Ros} (\citeyear{Ros}), in that we do not
use a lexicon and we do not restrict the domain of the noun-modifier pairs.

In work that is related to \citeauthor{Dol} (\citeyear{Dol}) (see Section~\ref{subsec:metana}),
\citeauthor{Van} (\citeyear{Van}) used hand-built rules,
together with the LKB derived from LDOCE, to classify noun-modifier pairs. Tested
with 97 pairs extracted from the Brown corpus, the rules had an accuracy of 52\%.

\citeauthor{BarSzp} (\citeyear{BarSzp}) used memory based learning (MBL) for classifying
semantic relations. The memory base stored triples, consisting of a noun, its modifier,
and (if available) a marker. The marker was either a preposition or an appositive marker
when the noun-modifier pair was found in text next to a preposition or an apposition. A
new noun-modifier pair was classified by looking for the nearest neighbours in the
memory base. The distance (similarity) measure was based on literal matches between
the elements in the triples, which constrained the algorithm's ability to generalize from
past examples.

Some research has concentrated on learning particular semantic relations, such as {\em part\_of}
\cite{BerCha} or {\em type\_of} \cite{Hea}. These are specific instances of
the more general problem considered here (see Table~\ref{classes}).

The algorithm of \citeauthor{LapLas} (\citeyear{LapLas}) for paraphrasing
logical metonymy can be viewed as a method for making semantic relations
explicit. Some of the logical metonymies they consider take the form of
noun-modifier pairs, such as ``difficult language'', which can be non-metonymically
paraphrased as ``language that is difficult to learn''. However, most noun-modifier
pairs are not logical metonymies, and the two tasks seem different, since
it is difficult to cast logical metonymy as a classification problem.

In this paper, we apply a measure of analogical similarity to classifying noun-modifier
relations, but, in principle, this could work the other way around; an algorithm for
classifying noun-modifier relations could be used to solve SAT-style verbal analogy
problems. The stem pair and each of the choice pairs could be classified according to
their semantic relations. Ideally, the stem and the correct choice would be classified as
having the same semantic relation, whereas the incorrect choices would have different
semantic relations. We have done some preliminary experiments with this approach, but
have not yet had any success.

\subsection{Web Mining}\label{subsec:webmining}

Our learning algorithm relies on a very large corpus of web pages. We obtain
information about the frequency of various patterns of words by querying
a web search engine (AltaVista). Other researchers have used web search
engines to acquire data for natural language processing applications. For
example, \citeauthor{Res99a} (\citeyear{Res99a}) used AltaVista to find bilingual text.
Our approach is different in that it only needs frequency information and not the text itself;
the only information we use from AltaVista is the {\em hit count} (the number of web 
pages that match the given query).

The use of hit counts from web search engines to obtain lexical statistical information
was introduced by \citeauthor{Tur} (\citeyear{Tur}), who used hit counts from AltaVista
to estimate Pointwise Mutual Information (PMI). This approach to estimating
PMI resulted in a good measure of {\em semantic similarity} between pairs of words. When evaluated 
with multiple-choice synonym questions, taken from the Test of English as a Foreign Language
(TOEFL), the PMI estimate achieved a score of 73.75\% \cite{Tur}. In comparison, the average 
human TOEFL score was 64.5\%.

\citeauthor{TurLit} (\citeyear{TurLit}) used AltaVista hit counts to determine
the {\em semantic orientation} of words. A word has a positive semantic orientation
when it conveys praise (honest, cute) and a negative orientation when it indicates
criticism (horrible, cruel). Semantic orientation varies in both direction (positive or negative) 
and degree (mild to strong). The algorithm was experimentally tested with 3,596 words (including 
adjectives, adverbs, nouns, and verbs) that were manually labeled positive (1,614 words) 
and negative (1,982 words). It attained an accuracy of 82.8\% on the full test set, but 
the accuracy was greater than 95\% when the algorithm was allowed to abstain from classifying mild 
words. 

In this paper, we use hit counts to measure the similarity between semantic relations,
rather than the similarity between individual concepts \cite{Tur}. The above papers
share the idea of using web search engines to exploit a huge corpus for natural language 
processing applications, but the details of the applications are quite different.

\section{Solving Verbal Analogy Problems}\label{sec:solving}

In Section~\ref{subsec:anapro}, we examine the task of solving verbal analogies. 
Section~\ref{subsec:vsmapproach} outlines the
application of the Vector Space Model to this task. The experimental results are presented
in Section~\ref{subsec:anaexper} and discussed in Section~\ref{subsec:anadiscuss}.

\subsection{Analogy Problems}\label{subsec:anapro}

The semantic relation between a pair of words may have no direct, obvious connection to
the individual words themselves. In an analogy {\em A:B::C:D\/}, there is not necessarily much
in common between $A$ and $C$ or between $B$ and $D$. Consider the analogy
``traffic:street::water:riverbed'' (one of our SAT questions). Traffic flows down a street;
water flows down a riverbed. A street carries traffic; a riverbed carries water. This
analogy is not superficial; there is a relatively large body of work on the mathematics of
hydrodynamics applied to modeling automobile traffic flow \cite{Dag, Zha, Yietal}.
Yet, if we look at the positions of these four words in the WordNet
hierarchy \cite{Fel}, it appears that they have little in common (see Table~\ref{hierarchy}).
``Traffic'' and ``water'' belong to different hierarchies (the former is a ``group'' and the
latter is a ``physical thing''). ``Street'' and ``riverbed'' are both ``physical objects'', but it
takes several steps up the hierarchy to find the abstract class to which they both belong.

\begin{table}[t]
\begin{tabular}{lll}
\hline
traffic  &  $\Rightarrow$  &
  collection $\Rightarrow$ group, grouping                                                             \\
water    &  $\Rightarrow$  &
  liquid $\Rightarrow$ fluid $\Rightarrow$ substance, matter $\Rightarrow$ entity, physical thing      \\
street   &  $\Rightarrow$  &
  thoroughfare $\Rightarrow$ road, route $\Rightarrow$ way $\Rightarrow$ artifact $\Rightarrow$        \\
         &                 &
  physical object $\Rightarrow$ entity, physical thing                                                 \\
riverbed &  $\Rightarrow$  &
  bed, bottom $\Rightarrow$ natural depression $\Rightarrow$ geological formation $\Rightarrow$        \\
         &                 &
  natural object $\Rightarrow$ physical object $\Rightarrow$ entity, physical thing                    \\
\hline
\end{tabular}
\caption[]{Location of the four words in the WordNet hierarchy.}\label{hierarchy}
\end{table}

This example illustrates that the similarity of semantic relations between words is not
directly reducible to the semantic similarity of individual words. Algorithms that have
been successful for individual words \cite{Les, ChuHan, Dun, Sma, Res, LanDum, Tur, PanLin}
will not work for semantic relations without significant modification.

\subsection{VSM Approach}\label{subsec:vsmapproach}

Given candidate analogies of the form {\em A:B::C:D\/}, we wish to assign scores to the
candidates and select the highest scoring candidate. The quality of a candidate analogy
depends on the similarity of the semantic relation $R_1$ between $A$ and $B$ to the semantic
relation $R_2$ between $C$ and $D$. The relations $R_1$ and $R_2$ are not given to us; the task is to
infer these relations automatically. Our approach to this task, inspired by the Vector
Space Model of information retrieval \cite{SalMcG, Sal}, is to
create vectors, $r_1$ and $r_2$, that represent features of $R_1$ and $R_2$, and then measure the
similarity of $R_1$ and $R_2$ by the cosine of the angle $\theta$ between $r_1$ and $r_2$:

\[
r_1  = \left\langle {r_{1,1} , \cdots ,r_{1,n} } \right\rangle,
\]

\[
r_2  = \left\langle {r_{2,1} , \cdots ,r_{2,n} } \right\rangle,
\]

\[
{\rm cosine(}\theta {\rm )
= }\frac{{\sum\limits_{i = 1}^n {r_{1,i}  \cdot r_{2,i} } }}
{{\sqrt {\sum\limits_{i = 1}^n {(r_{1,i} )^2  \cdot \sum\limits_{i = 1}^n {(r_{2,i} )^2 } } } }}
= \frac{{r_1  \cdot r_2 }}{{\sqrt {r_1  \cdot r_1 }  \cdot \sqrt {r_2  \cdot r_2 } }}
= \frac{{r_1  \cdot r_2 }}{{\left\| {r_1 } \right\| \cdot \left\| {r_2 } \right\|}}\;.
\]

We create a vector, $r$, to characterize the relationship between two words, $X$ and $Y$, by
counting the frequencies of various short phrases containing $X$ and $Y$. We use a list of 64
joining terms (see Table~\ref{joining}), such as ``of'', ``for'', and ``to'', to form 128 phrases
that contain $X$ and $Y$, such as ``$X$ of $Y$'', ``$Y$ of $X$'', ``$X$ for $Y$'',
``$Y$ for $X$'', ``$X$ to $Y$'', and ``$Y$ to $X$''. We
then use these phrases as queries for a search engine and record the number of hits
(matching documents) for each query. This process yields a vector of 128 numbers.

\def\tablesize{\fontsize{7}{8} \selectfont}

\begin{table}[t]
\begin{tablesize}
\begin{tabular}{rlrlrlrl}
\hline
1  & `` ''             & 17 & `` get* ''       & 33 & `` like the ''    & 49 & `` then ''      \\
2  & `` * not ''       & 18 & `` give* ''      & 34 & `` make* ''       & 50 & `` this ''      \\
3  & `` * very ''      & 19 & `` go ''         & 35 & `` need* ''       & 51 & `` to ''        \\
4  & `` after ''       & 20 & `` goes ''       & 36 & `` not ''         & 52 & `` to the ''    \\
5  & `` and not ''     & 21 & `` has ''        & 37 & `` not the ''     & 53 & `` turn* ''     \\
6  & `` are ''         & 22 & `` have ''       & 38 & `` of ''          & 54 & `` use* ''      \\
7  & `` at ''          & 23 & `` in ''         & 39 & `` of the ''      & 55 & `` when ''      \\
8  & `` at the ''      & 24 & `` in the ''     & 40 & `` on ''          & 56 & `` which ''     \\
9  & `` become* ''     & 25 & `` instead of '' & 41 & `` onto ''        & 57 & `` will ''      \\
10 & `` but not ''     & 26 & `` into ''       & 42 & `` or ''          & 58 & `` with ''      \\
11 & `` contain* ''    & 27 & `` is ''         & 43 & `` rather than '' & 59 & `` with the ''  \\
12 & `` for ''         & 28 & `` is * ''       & 44 & `` such as ''     & 60 & `` within ''    \\
13 & `` for example '' & 29 & `` is the ''     & 45 & `` than ''        & 61 & `` without ''   \\
14 & `` for the ''     & 30 & `` lack* ''      & 46 & `` that ''        & 62 & `` yet ''       \\
15 & `` from ''        & 31 & `` like ''       & 47 & `` the ''         & 63 & ``'s ''         \\
16 & `` from the ''    & 32 & `` like * ''     & 48 & `` their ''       & 64 & ``'s * ''       \\
\hline
\end{tabular}
\vspace{-5pt}
\end{tablesize}
\caption[]{The 64 joining terms.}\label{joining}
\end{table}

We have found that the accuracy of this approach to scoring analogies improves when we use
the logarithm of the frequency. That is, if $x$ is the number of hits for a query, then the
corresponding element in the vector $r$ is log($x$ + 1).\endnote{We add 1 to $x$ because the
logarithm of zero is undefined. The base of the logarithm does not matter, since all
logarithms are equivalent up to a constant multiplicative factor. Any constant
factor drops out when calculating the cosine.} \citeauthor{Rug} (\citeyear{Rug}) found that
using the logarithm of the frequency also yields better results when measuring the semantic
similarity of individual words, and log-based measures for similarity are used
in \citeauthor{Lin} (\citeyear{Lin}) and \citeauthor{Res99b} (\citeyear{Res99b}).
Logarithms are also commonly used in the VSM for information retrieval \cite{SalBuc}.

We used the AltaVista search engine (http://www.altavista.com/) in the following
experiments. At the time our experiments were done, we estimate that AltaVista's index
contained about 350 million English web pages (about $10^{11}$ words). We chose AltaVista
for its ``*'' operator, which serves two functions:

\begin{enumerate}
\item {\em Whole word matching:} In a quoted phrase, an asterisk can match any whole word.
The asterisk must not be the first or last character in the quoted phrase. The asterisk
must have a blank space immediately before and after it. For example, the query
``immaculate * very clean'' will match ``immaculate and very clean'', ``immaculate is
very clean'', ``immaculate but very clean'', and so on.
\item {\em Substring matching:} Embedded in a word, an asterisk can match zero to five
characters. The asterisk must be preceded by at least three regular alphabetic
characters. For example, ``colo*r'' matches ``color'' and ``colour''.
\end{enumerate}

Some of the joining terms in Table~\ref{joining} contain an asterisk, and we also use the asterisk for
stemming, as specified in Table~\ref{stemming}. For instance, consider the pair ``restrained:limit'' and
the joining term `` * very ''. Since ``restrained'' is ten characters long, it is stemmed to
``restrai*''. Since ``limit'' is five characters long, it is stemmed to ``limit*''. Joining these
stemmed words, we have the two queries ``restrai* * very limit*'' and ``limit* * very
restrai*''. The first query would match ``restrained and very limited'', ``restraints are very
limiting'', and so on. The second query would match ``limit is very restraining'', ``limiting
and very restraining'', and so on.

\begin{table}[t]
\begin{tabular}{ll}
\hline
Stemming rule                                 & Example                        \\
\hline
If 10 $<$ length, then replace the            & advertisement $\to$ advertise* \\
~~~last 4 characters with ``*''.              &                                \\
If 8 $<$ length $\le$ 10, then replace        & compliance $\to$ complia*      \\
~~~the last 3 characters with ``*''.          &                                \\
If 2 $<$ length $\le$ 8, then append          & rhythm $\to$ rhythm*           \\
~~~``*'' to the end.                          &                                \\
If length $\le$ 2, then do nothing.           & up $\to$ up                    \\
\hline
\end{tabular}
\vspace{-15pt}
\caption[]{Stemming rules.}\label{stemming}
\end{table}

The vector $r$ is a kind of {\em signature} of the semantic relationship between $X$ and $Y$.
Consider the analogy traffic:street::water:riverbed. The words ``traffic'' and
``street'' tend to appear together in phrases such as ``traffic in the street'' (544 hits on
AltaVista) and ``street with traffic'' (460 hits), but not in phrases such as ``street on traffic''
(7 hits) or ``street is traffic'' (15 hits). Similarly, ``water'' and ``riverbed'' may appear
together as ``water in the riverbed'' (77 hits), but ``riverbed on water'' (0 hits) would be
unlikely. Therefore the angle $\theta$ between the vector $r_1$ for traffic:street and the vector $r_2$ for
water:riverbed tends to be relatively small, and hence cosine($\theta$) is relatively large.

To answer an SAT analogy question, we calculate the cosines of the angles between the
vector for the stem pair and each of the vectors for the choice pairs. The algorithm
guesses that the answer is the choice pair with the highest cosine. This 
learning algorithm makes no use of labeled training data.

The joining terms in Table~\ref{joining} are similar to the patterns used by 
\citeauthor{Hea} (\citeyear{Hea}) and \citeauthor{BerCha} (\citeyear{BerCha}).
\citeauthor{Hea} (\citeyear{Hea}) used various patterns to discover hyponyms in
a large corpus. For example, the pattern ``$N\!P_0$ such as $N\!P_1$''
provides evidence that $N\!P_1$ is a hyponym of $N\!P_0$. Thus the phrase 
``the bow lute, such as the Bambara ndang'' suggests that the {\em Bambara ndang} 
is a type of (hyponym of) {\em bow lute} \cite{Hea}. The joining term ``such as'' 
is item 44 in Table~\ref{joining}.

\citeauthor{BerCha} (\citeyear{BerCha}) used patterns to discover meronyms
in a large corpus. The pattern ``$N\!P_0$ of the $N\!P_1$'' suggests that $N\!P_0$
may be a part of (meronym of) $N\!P_1$ (``the basement of the building'') \cite{BerCha}. The joining 
term ``of the'' is item 39 in Table~\ref{joining}. 

Our work is different from
\citeauthor{Hea} (\citeyear{Hea}) and \citeauthor{BerCha} (\citeyear{BerCha})
in that they only consider a single semantic relation, rather than multiple classes of semantic 
relations. Also, we are using these patterns to generate features in
a high-dimensional vector, rather than using them to search for particular instances
of a specific semantic relationship.

\subsection{Experiments}\label{subsec:anaexper}

In the following experiments, we evaluate the VSM approach to solving analogies using a
set of 374 SAT-style verbal analogy problems. This is the same set of questions as was
used in \citeauthor{Turetal} (\citeyear{Turetal}), but the experimental setup is different. The
ensemble merging rules of \citeauthor{Turetal} (\citeyear{Turetal}) use supervised learning, so
the 374 questions were separated there into 274 training questions and 100 testing questions.
However, the VSM approach by itself needs no labeled training data, so we are able to test it
here on the full set of 374 questions.

Section~\ref{subsubsec:recognizing} considers the task of {\em recognizing} analogies and 
Section~\ref{subsubsec:generating} takes a step
towards {\em generating} analogies.

\subsubsection{Recognizing Analogies}\label{subsubsec:recognizing}

Following standard practice in information retrieval \cite{Rij}, we define
precision, recall, and $F$ as follows:

\[
{\rm precision } = \frac{{{\rm number~of~correct~guesses}}}{{{\rm total~number~of~guesses~made}}}
\]

\[
{\rm recall } = \frac{{{\rm number~of~correct~guesses}}}{{{\rm maximum~possible~number~correct}}}
\]

\[
F = \frac{{2 \times {\rm precision} \times {\rm recall}}}{{{\rm precision + recall}}}\;.
\]

\noindent When any of the denominators are zero, we define the result to be zero. All three 
of these performance measures range from 0 to 1, and larger values are better than smaller values.

Table~\ref{results1} shows the experimental results for our set of 374 analogy questions. Five
questions were skipped because the vector for the stem pair was entirely zeros. Since
there are five choices for each question, random guessing would yield a recall of 20\%.
The algorithm is clearly performing much better than random guessing ($p < 0.0001$ according 
to Fisher's Exact test). 

\begin{table}[t]
\begin{tabular}{lrr}
\hline
            & Number      & Percent   \\
\hline
Correct     & 176         & 47.1\%    \\
Incorrect   & 193         & 51.6\%    \\
Skipped     & 5           & 1.3\%     \\
Total       & 374         & 100.0\%   \\
\hline
Precision   & 176 / 369   & 47.7\%    \\
Recall      & 176 / 374   & 47.1\%    \\
$F$         &             & 47.4\%    \\
\hline
\end{tabular}
\vspace{-5pt}
\caption[]{Results of experiments with the 374 analogy questions.}\label{results1}
\end{table}

Our analogy question set \cite{Turetal} was constructed from books and web sites
intended for students preparing for the SAT college entrance exam, including 90
questions from unofficial SAT preparation web sites, 14 questions from the Educational
Testing Service (ETS) web site (http://www.ets.org/), 190 questions scanned in from a
book with actual SAT exams (Claman, 2000), and 80 questions typed from SAT
guidebooks.

The SAT I test consists of 78 verbal questions and 60 math questions (there is also an SAT
II test, covering specific subjects, such as chemistry). The questions are multiple choice,
with five choices per question. The verbal and math scores are reported separately. The
raw SAT I score is calculated by giving one point for each correct answer, zero points for
skipped questions, and subtracting one quarter point for each incorrect answer. The
quarter point penalty for incorrect answers is chosen so that the expected raw score for
random guessing is zero points. The raw score is then converted to a scaled score that
ranges from 200 to
800.\endnote{See http://www.collegeboard.com/prod\_downloads/about/news\_info/cbsenior/ \\
yr2002/pdf/two.pdf.}
The College Board publishes information about the percentile
rank of college-bound senior high school students for the SAT~I verbal and math
questions.\endnote{See http://www.collegeboard.com/prod\_downloads/about/news\_info/cbsenior/ \\
yr2002/pdf/threeA.pdf.}
On the verbal SAT test, the mean scaled score for 2002 was 504. We used
information from the College Board to make Table~\ref{sat}.

Analogy questions are only a subset of the 78 verbal SAT questions. If we assume that
the difficulty of our 374 analogy questions is comparable to the difficulty of other verbal
SAT questions, then we can estimate that the average college-bound senior would
correctly answer about 57\% of the 374 analogy questions. We can also estimate that the
performance of the VSM approach corresponds to a percentile rank of 29$\pm$3.
\citeauthor{Cla} (\citeyear{Cla}) suggests that the analogy questions may be
somewhat harder than other verbal SAT
questions, so we may be slightly overestimating the mean human score on the analogy
questions.

\begin{table}[t]
\begin{tabular}{lrrrr}
\hline
Note                        & Percent       & SAT I      & SAT I        & Percentile     \\
                            & correct       & raw score  & scaled score & rank           \\
                            & (no skipping) & verbal     & verbal       &                \\
\hline
                            & 100\%         & 78         & 800$\pm$10   & 100.0$\pm$0.5  \\
                            & 92\%          & 70         & 740$\pm$20   & 98.0$\pm$1.0   \\
                            & 82\%          & 60         & 645$\pm$15   & 88.5$\pm$2.5   \\
                            & 71\%          & 50         & 580$\pm$10   & 74.0$\pm$3.0   \\
College-bound mean --       & 57\%          & 36         & 504$\pm$10   & 48.0$\pm$3.5   \\
VSM algorithm --            & 47\%          & 26         & 445$\pm$10   & 29.0$\pm$3.0   \\
                            & 41\%          & 20         & 410$\pm$10   & 18.5$\pm$2.5   \\
                            & 30\%          & 10         & 335$\pm$15   & 5.5$\pm$1.5    \\
Random guessing --          & 20\%          & 0          & 225$\pm$25   & 0.5$\pm$0.5    \\
\hline
\end{tabular}
\vspace{-20pt}
\caption[]{Verbal SAT scores.}\label{sat}
\end{table}

There is a well-known trade-off between precision and recall: By skipping hard
questions, we can increase precision at the cost of decreased recall. By making multiple
guesses for each question, we can increase recall at the cost of decreased precision. The $F$
measure is the harmonic mean of precision and recall. It tends to be largest when
precision and recall are balanced.

For some applications, precision may be more important than recall, or vice versa. Thus it
is useful to have a way of adjusting the balance between precision and recall. We
observed that the difference between the cosine of the best choice and the cosine of the
second best choice (the largest cosine minus the second largest) seems to be a good
indicator of whether the guess is correct. We call this difference the {\em margin}. By setting a
threshold on the margin, we can trade off precision and recall.

When the threshold on the margin is a positive number, we skip every question for which
the margin is less than the threshold. This tends to increase precision and decrease recall.
On the other hand, when the threshold on the margin is negative, we make two guesses
(both the best and the second best choices) for every question for which the margin is less
than the absolute value of the threshold. Ties are unlikely, but if they happen, we break
them randomly.

Consider the example in Table~\ref{example1}. The best choice is (e) and the second best choice is (c).
(In this case, the best choice is correct.) The margin is $0.00508$ ($0.69265$ minus $0.68757$).
If the threshold is between $-0.00508$ and $+0.00508$, then the output is choice (e) alone. If
the threshold is greater than $+0.00508$, then the question is skipped. If the threshold is
less than $-0.00508$, then the output is both (e) and~(c).

\begin{table}[t]
\begin{tabular}{lllr}
\hline
Stem pair:     &     & traffic:street   & Cosine  \\
\hline
Choices:       & (a) & ship:gangplank   & 0.31874 \\
               & (b) & crop:harvest     & 0.57234 \\
               & (c) & car:garage       & 0.68757 \\
               & (d) & pedestrians:feet & 0.49725 \\
               & (e) & water:riverbed   & 0.69265 \\
\hline
\end{tabular}
\caption[]{An example of an analogy question, taken from the set of 374 questions.}\label{example1}
\end{table}

\begin{figure}[t]
\centerline{\includegraphics[width=\columnwidth,keepaspectratio]{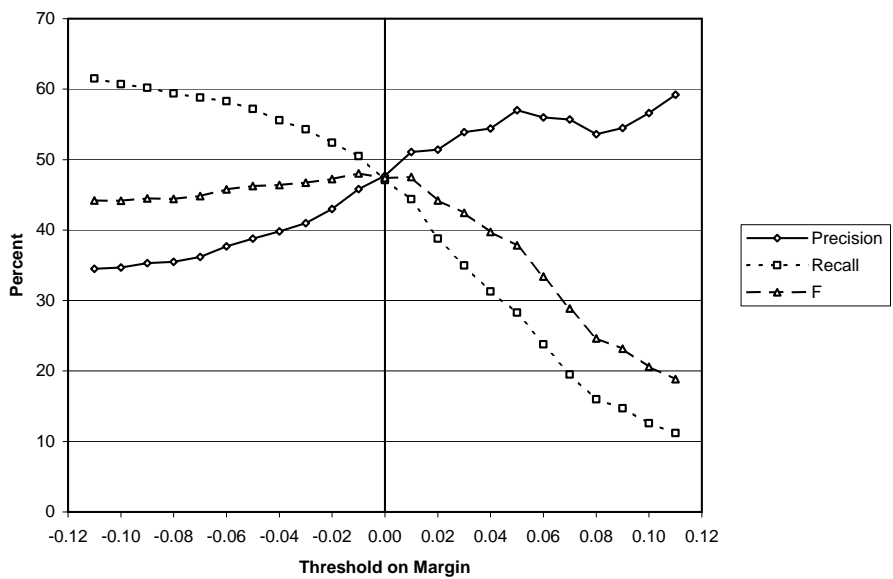}}
\caption{Precision and recall for 374 SAT-style analogy questions.}\label{figure1}
\end{figure}

Figure~\ref{figure1} shows precision, recall, and $F$ as the threshold on the margin varies from $-0.11$
to $+0.11$. The vertical line at the threshold zero corresponds to the situation in Table~\ref{results1}.
With a threshold of $+0.11$, precision reaches 59.2\% and recall drops to 11.2\%. With a
threshold of $-0.11$, recall reaches 61.5\% and precision drops to 34.5\%. These
precision-recall results compare favourably with typical results in information retrieval
\cite{VooHar}.

In Figure~\ref{figure1}, we see that the $F$ value reaches its maximum when the threshold on the
margin is near zero. This is expected, since $F$ is intended to favour a balance between
precision and recall.

The experiments presented here required 287,232 queries to AltaVista (374 analogy
questions $\times$ 6 word pairs per question $\times$ 128 queries per word pair). Although
AltaVista is willing to support automated queries of the kind described here, as a courtesy, we
inserted a five second delay between each query. Thus processing the 287,232 queries
took about seventeen days.

\subsubsection{Generating Analogies}\label{subsubsec:generating}

The results so far suggest that our algorithm is capable of {\em recognizing} analogies with
some degree of success, but an interesting question is whether it might be capable of
{\em generating} analogies. That is, given a stem pair, the algorithm can often pick out the
correct choice pair from a set of five choices, but generating a verbal analogy from
scratch is a more difficult problem. One approach to the generation problem is to try to
reduce it to the recognition problem, by randomly generating candidate analogies and
then trying to recognize good analogies among the candidates.

As a first step towards generating analogies, we expanded the number of choices for each
stem pair. We dropped the five questions for which the stem vector was all zeros, leaving
369 questions. For each of the remaining questions, we combined the 369 correct choice
pairs. For each of the 369 stem pairs, the algorithm had to choose the correct word pair
from among the 369 possible answers.

For each of the 369 stem pairs, the 369 choice pairs were sorted in order of decreasing
cosine. We then examined the top ten most highly ranked choices to see whether the
correct choice was among them. Table~\ref{results2} shows the result of this experiment. The first
row in the table shows that the first choice was correct for 31 of the 369 stems (8.4\%).
The last row shows that the correct choice appears somewhere among the top ten choices
29.5\% of the time. With random guessing, the correct choice would appear among the top
ten 2.7\% of the time (10/369 = 0.027).

\begin{table}[t]
\begin{tabular}{rrrrr}
\hline
Rank  & Matches & Matches & Cumulative & Cumulative \\
\#    & \#      & \%      & \#         & \%         \\
\hline
1     & 31      & 8.4\%   & 31         & 8.4\%      \\
2     & 19      & 5.1\%   & 50         & 13.6\%     \\
3     & 13      & 3.5\%   & 63         & 17.1\%     \\
4     & 11      & 3.0\%   & 74         & 20.1\%     \\
5     & 6       & 1.6\%   & 80         & 21.7\%     \\
6     & 7       & 1.9\%   & 87         & 23.6\%     \\
7     & 9       & 2.4\%   & 96         & 26.0\%     \\
8     & 5       & 1.4\%   & 101        & 27.4\%     \\
9     & 5       & 1.4\%   & 106        & 28.7\%     \\
10    & 3       & 0.8\%   & 109        & 29.5\%     \\
\hline
\end{tabular}
\vspace{-5pt}
\caption[]{Selecting the correct word pair from a set of 369 choices.}\label{results2}
\end{table}

\begin{table}[ht]
\begin{tabular}{rlrr}
\hline
Rank      & Word pair              & Cosine        & Question number \\
\hline
Stem      & barley:grain           &               & 33              \\
\hline
1         & aluminum:metal         & 0.8928        & 198             \\
2         & beagle:dog             & 0.8458        & 190             \\
{\em 3}   & {\em pine:tree}        & {\em 0.8451}  & {\em 33}        \\
4         & emerald:gem            & 0.8424        & 215             \\
5         & sugar:sweet            & 0.8240        & 327             \\
6         & pseudonym:name         & 0.8151        & 240             \\
7         & mile:distance          & 0.8142        & 21              \\
8         & oil:lubricate          & 0.8133        & 313             \\
9         & novel:book             & 0.8117        & 182             \\
10        & minnow:fish            & 0.8111        & 193             \\
\hline
Stem      & tourniquet:bleeding    &               & 46              \\
\hline
1         & antidote:poisoning     & 0.7540        & 308             \\
2         & belligerent:fight      & 0.7482        & 84              \\
3         & chair:furniture        & 0.7481        & 107             \\
4         & mural:wall             & 0.7430        & 302             \\
5         & reciprocate:favor      & 0.7429        & 151             \\
6         & menu:diner             & 0.7421        & 284             \\
7         & assurance:uncertainty  & 0.7287        & 8               \\
8         & beagle:dog             & 0.7210        & 19              \\
9         & canvas:painting        & 0.7205        & 5               \\
10        & ewe:sheep              & 0.7148        & 261             \\
\hline
\end{tabular}
\vspace{-10pt}
\caption[]{Two examples of stem pairs and the top ten choices.}\label{example2}
\end{table}

This experiment actually underestimates the quality of the output. Table~\ref{example2} shows the top
ten choices for two stem pairs. For the first stem pair, barley:grain, the correct choice,
according to the original formulation of the test, is pine:tree, which is the third choice
here. The semantic relation between barley and grain is {\em type\_of} (hyponym), so the first
two choices, aluminum:metal and beagle:dog, are perfectly acceptable alternatives. In
fact, it could be argued that aluminum:metal is a better choice, because aluminum and
barley are mass nouns (i.e., they do not form plurals), but pine is a count noun (e.g., ``I
have two pines in my yard.'').

For the second stem pair in Table~\ref{example2}, tourniquet:bleeding, the original correct choice,
splint:movement, is not among the top ten choices. (A tourniquet prevents or
reduces bleeding; a splint prevents or reduces movement.) However, the first choice,
antidote:poisoning, is a good alternative. (A tourniquet is used to treat bleeding; an
antidote is used to treat poisoning.) The seventh choice, assurance:uncertainty, also seems
reasonable. (Assurance puts an end to uncertainty; a tourniquet puts an end to 
bleeding.)\endnote{Even if the reader does not agree with our judgments about what 
``seems reasonable'', the performance of 29.5\% remains valid as a
lower bound on the quality of the output; we only disagree on how far the quality
is from this lower bound.}

\subsection{Discussion}\label{subsec:anadiscuss}

As mentioned in Section~\ref{subsec:metana}, the VSM algorithm performs as well as an ensemble of
twelve other modules \cite{Turetal}. All of the other modules employed various
lexical resources (WordNet, Dictionary.com, and Wordsmyth.net), whereas the VSM
module learns from a large corpus of unlabeled text, without a lexicon. The VSM
performance of 47.1\% correct is well above the 20\% correct that would be expected for
random guessing, but it is also less than the 57\% correct that would be expected for the
average college-bound senior high school student.

When the number of choices for each stem is expanded from five to 369, the correct
choice is among the top ten choices 29.5\% of the time, where random guessing would
give 2.7\%. There is certainly much room for improvement, but there is also good
evidence that verbal analogies can be solved algorithmically.

The list of joining terms in Table~\ref{joining} is somewhat arbitrary. This list was based on
preliminary experiments with a development set of analogy questions. The terms in the
list were selected by intuition and there is no reason to
believe they are optimal. It might be possible to automatically learn joining terms,
perhaps by extending the algorithm of \citeauthor{RavHov} (\citeyear{RavHov}).

We attempted to take a more principled approach to the
joining terms, by creating a larger list of 142 joining terms, and then using feature
selection algorithms (forward selection, backward elimination, genetic algorithm selection)
to select an optimal subset of the features. None of the selected subsets were able to
achieve statistically significantly better performance in cross-validation testing compared
to the original set in Table~\ref{joining}. The subsets seemed to overfit the training questions. We
believe that this problem can be fixed with a larger set of questions. 

The idea of using the {\em margin} to trade off precision and recall was inspired
by Support Vector Machines, which use a somewhat related concept
of margin \cite{CriSha}. This suggests the possibility of using a supervised learning
approach, in which a training set would be used to tune parameters to maximize
the margin. We believe that this is a good approach, but so far we have not been 
successful with it. 

The execution time (seventeen days) would be much less if we had a local copy of the
AltaVista database. Progress in hardware will soon make it practical for a standard
desktop computer to search in a local copy of a corpus of this size (about $10^{11}$ words).

\section{Noun-Modifier Relations}\label{sec:nounmods}

In Section~\ref{subsec:nounmodapps}, we discuss applications for an algorithm for
classifying noun-modifier relations. 
Section~\ref{subsec:classes} presents the classes of noun-modifier relations that are used in our
experiments \cite{NasSzp}. The classification algorithm is described in
Section~\ref{subsec:nearest}. The experiments are in Section~\ref{subsec:nounmodexper}, 
followed by discussion of the results in Section~\ref{subsec:nounmoddiscuss}.

\subsection{Applications}\label{subsec:nounmodapps}

Noun-modifier word pairs are common in English and other languages. An algorithm
for classification of noun-modifier relations would be useful in machine translation,
information extraction, and word sense disambiguation. We illustrate this with
examples taken from the collection of 600 labeled noun-modifier pairs used in
our experiments (see Table~\ref{classes}).

{\em Machine translation}: A noun-modifier pair such as ``electron microscope'' might not have
a direct translation into an equivalent noun-modifier pair in another language. In the
translation process, it may be necessary to expand the noun-modifier pair into a longer
phrase, explicitly stating the implicit semantic relation. Is the semantic relation {\em purpose}
(a microscope for electrons; e.g., for viewing electrons), {\em instrument} (a microscope 
that uses electrons), or {\em material} (a microscope made out of electrons)? The 
answer to this question may be used in translation. (The terms
{\em purpose}, {\em instrument}, and {\em material} are explained in Table~\ref{classes}.)

{\em Information extraction}: A typical information extraction task would be to process news
stories for information about wars. The task may require finding information about the
parties involved in the conflict. It would be important to know that the semantic relation
in the noun-modifier pair ``cigarette war'' is {\em topic} (a war about cigarettes), not {\em agent}
(a war by cigarettes; i.e., cigarettes are fighting the war).

{\em Word sense disambiguation}: The word ``plant'' might refer to an industrial plant or a
living organism. If a document contains the noun-modifier pair ``plant food'', a word
sense disambiguation algorithm can take advantage of the information that the semantic
relation involved is {\em beneficiary} (the plant benefits from the food), rather than {\em source} (the
plant is the source of the food).

\subsection{Classes of Relations}\label{subsec:classes}

The following experiments use the 600 labeled noun-modifier pairs of \citeauthor{NasSzp}
(\citeyear{NasSzp}). This data set includes information about the part of speech and
WordNet synset (synonym set; i.e., word sense tag) of each word, but our algorithm does
not use this information.

Table~\ref{classes} lists the 30 classes of semantic relations. The table is based on Appendix A of
\citeauthor{NasSzp} (\citeyear{NasSzp}), with some simplifications. The original table listed
several semantic relations for which there were no instances in the data set. These were
relations that are typically expressed with longer phrases (three or more words), rather
than noun-modifier word pairs. For clarity, we decided not to include these relations in
Table~\ref{classes}.

\def\tablesize{\fontsize{7}{8} \selectfont}

\begin{table}[ht]
\begin{tablesize}
\begin{sloppypar}
\begin{tabular}{llll}
\hline
Relation          & Abbr.     & Example phrase           & Description                                 \\
\hline
{\sc Causality}   &           &                          &                                             \\
\hline
cause             & cs        & flu virus (*)            & $H$ makes $M$ occur or exist, $H$ is        \\
                  &           &                          & ~~necessary and sufficient                  \\
effect            & eff       & exam anxiety             & $M$ makes $H$ occur or exist, $M$ is        \\
                  &           &                          & ~~necessary and sufficient                  \\
purpose           & prp       & concert hall ($\dagger$) & $H$ is for {\em V}-ing $M$, $M$ does not    \\
                  &           &                          & ~~necessarily occur or exist                \\
detraction        & detr      & headache pill            & $H$ opposes $M$, $H$ is not sufficient      \\
                  &           &                          & ~~to prevent $M$                            \\
\hline
{\sc Temporality} &           &                          &                                             \\
\hline
frequency         & freq      & daily exercise           & $H$ occurs every time $M$ occurs            \\
time at           & tat       & morning exercise         & $H$ occurs when $M$ occurs                  \\
time through      & tthr      & six\mbox{-}hour meeting  & $H$ existed while $M$ existed, $M$ is       \\
                  &           &                          & ~~an interval of time                       \\
\hline
{\sc Spatial}     &           &                          &                                             \\
\hline
direction         & dir       & outgoing mail            & $H$ is directed towards $M$, $M$ is         \\
                  &           &                          & ~~not the final point                       \\
location          & loc       & home town                & $H$ is the location of $M$                  \\
location at       & lat       & desert storm             & $H$ is located at $M$                       \\
location from     & lfr       & foreign capital          & $H$ originates at $M$                       \\
\hline
{\sc Participant} &           &                          &                                             \\
\hline
agent             & ag        & student protest          & $M$ performs $H$, $M$ is animate or         \\
                  &           &                          & ~~natural phenomenon                        \\
beneficiary       & ben       & student discount         & $M$ benefits from $H$                       \\
instrument        & inst      & laser printer            & $H$ uses $M$                                \\
object            & obj       & metal separator          & $M$ is acted upon by $H$                    \\
object property   & obj\_prop & sunken ship              & $H$ underwent $M$                           \\
part              & part      & printer tray             & $H$ is part of $M$                          \\
possessor         & posr      & national debt            & $M$ has $H$                                 \\
property          & prop      & blue book                & $H$ is $M$                                  \\
product           & prod      & plum tree                & $H$ produces $M$                            \\
source            & src       & olive oil                & $M$ is the source of $H$                    \\
stative           & st        & sleeping dog             & $H$ is in a state of $M$                    \\
whole             & whl       & daisy chain              & $M$ is part of $H$                          \\
\hline
{\sc Quality}     &           &                          &                                             \\
\hline
container         & cntr      & film music               & $M$ contains $H$                            \\
content           & cont      & apple cake               & $M$ is contained in $H$                     \\
equative          & eq        & player coach             & $H$ is also $M$                             \\
material          & mat       & brick house              & $H$ is made of $M$                          \\
measure           & meas      & expensive book           & $M$ is a measure of $H$                     \\
topic             & top       & weather report           & $H$ is concerned with $M$                   \\
type              & type      & oak tree                 & $M$ is a type of $H$                        \\
\hline
\end{tabular}
\vspace{-30pt}
\end{sloppypar}
\end{tablesize}
\caption[]{Classes of semantic relations \cite{NasSzp}.}\label{classes}
\end{table}

In this table, $H$ represents the head noun and $M$ represents the modifier. For example, in
``flu virus'', the head noun ($H$) is ``virus'' and the modifier ($M$) is ``flu'' (*). In English,
the modifier (typically a noun or adjective) usually precedes the head noun. In the
description of {\em purpose}, $V$ represents an arbitrary verb. In ``concert hall'', the hall is
for presenting concerts ($V$ is ``present'') or holding concerts ($V$ is ``hold'') ($\dagger$).

\citeauthor{NasSzp} (\citeyear{NasSzp})
organized the relations into groups. The five capitalized terms
in the ``Relation'' column of Table~\ref{classes} are the names of five groups of semantic relations.
(The original table had a sixth group, but there are no examples of this group in the data
set.) We make use of this grouping in Section~\ref{subsubsec:five}.

\clearpage

\subsection{Nearest-Neighbour Approach}\label{subsec:nearest}

The following experiments use single nearest-neighbour classification with leave-one-out
cross-validation. A vector of 128 numbers is calculated for each noun-modifier pair, as
described in Section~\ref{subsec:vsmapproach}. The similarity of two vectors is measured by the cosine of their
angle. For leave-one-out cross-validation, the testing set consists of a single vector and
the training set consists of the 599 remaining vectors. The data set is split 600 times, so
that each vector gets a turn as the testing vector. The predicted class of the testing vector
is the class of the single nearest neighbour (the vector with the largest cosine) in the
training set.

\subsection{Experiments}\label{subsec:nounmodexper}

Section~\ref{subsubsec:thirty} looks at the problem of assigning the 600 noun-modifier pairs to thirty
different classes. Section~\ref{subsubsec:five} considers the easier problem of assigning them to five
different classes.

\subsubsection{Thirty Classes}\label{subsubsec:thirty}

Table~\ref{results3} gives the precision, recall, and $F$ values for each of the 30 classes. The column
labeled ``class percent'' corresponds to the expected precision, recall, and $F$ for the simple
strategy of guessing each class randomly, with a probability proportional to the class size.
The actual average $F$ of 26.5\% is much larger than the average $F$ of 3.3\% that would be
expected for random guessing. The difference (23.2\%) is significant with 99\%
confidence ($p < 0.0001$, according to the paired t-test). The accuracy is 27.8\%
(167/600). 

\def\tablesize{\fontsize{8}{10} \selectfont}

\begin{table}[t]
\begin{tablesize}
\begin{sloppypar}
\begin{tabular}{lrrrrr}
\hline
Class name      & Class size & Class percent & Precision & Recall   & $F$      \\
\hline
ag              & 36         & 6.0\%         & 40.7\%    & 30.6\%   & 34.9\%   \\
ben             & 9          & 1.5\%         & 20.0\%    & 22.2\%   & 21.1\%   \\
cntr            & 3          & 0.5\%         & 40.0\%    & 66.7\%   & 50.0\%   \\
cont            & 15         & 2.5\%         & 23.5\%    & 26.7\%   & 25.0\%   \\
cs              & 17         & 2.8\%         & 18.2\%    & 11.8\%   & 14.3\%   \\
detr            & 4          & 0.7\%         & 50.0\%    & 50.0\%   & 50.0\%   \\
dir             & 8          & 1.3\%         & 33.3\%    & 12.5\%   & 18.2\%   \\
eff             & 34         & 5.7\%         & 13.5\%    & 14.7\%   & 14.1\%   \\
eq              & 5          & 0.8\%         & 0.0\%     & 0.0\%    & 0.0\%    \\
freq            & 16         & 2.7\%         & 47.1\%    & 50.0\%   & 48.5\%   \\
inst            & 35         & 5.8\%         & 15.6\%    & 14.3\%   & 14.9\%   \\
lat             & 22         & 3.7\%         & 14.3\%    & 13.6\%   & 14.0\%   \\
lfr             & 21         & 3.5\%         & 8.0\%     & 9.5\%    & 8.7\%    \\
loc             & 5          & 0.8\%         & 0.0\%     & 0.0\%    & 0.0\%    \\
mat             & 32         & 5.3\%         & 34.3\%    & 37.5\%   & 35.8\%   \\
meas            & 30         & 5.0\%         & 69.2\%    & 60.0\%   & 64.3\%   \\
obj             & 33         & 5.5\%         & 21.6\%    & 24.2\%   & 22.9\%   \\
obj\_prop       & 15         & 2.5\%         & 71.4\%    & 33.3\%   & 45.5\%   \\
part            & 9          & 1.5\%         & 16.7\%    & 22.2\%   & 19.0\%   \\
posr            & 30         & 5.0\%         & 23.5\%    & 26.7\%   & 25.0\%   \\
prod            & 16         & 2.7\%         & 14.7\%    & 31.3\%   & 20.0\%   \\
prop            & 49         & 8.2\%         & 55.2\%    & 32.7\%   & 41.0\%   \\
prp             & 31         & 5.2\%         & 14.9\%    & 22.6\%   & 17.9\%   \\
src             & 12         & 2.0\%         & 33.3\%    & 25.0\%   & 28.6\%   \\
st              & 9          & 1.5\%         & 0.0\%     & 0.0\%    & 0.0\%    \\
tat             & 30         & 5.0\%         & 64.3\%    & 60.0\%   & 62.1\%   \\
top             & 45         & 7.5\%         & 20.0\%    & 20.0\%   & 20.0\%   \\
tthr            & 6          & 1.0\%         & 40.0\%    & 33.3\%   & 36.4\%   \\
type            & 16         & 2.7\%         & 26.1\%    & 37.5\%   & 30.8\%   \\
whl             & 7          & 1.2\%         & 8.3\%     & 14.3\%   & 10.5\%   \\
\hline
Average         & 20         & 3.3\%         & 27.9\%    & 26.8\%   & 26.5\%   \\
\hline
\end{tabular}
\vspace{-10pt}
\end{sloppypar}
\end{tablesize}
\caption[]{The precision, recall, and $F$ for each of the 30 classes of semantic relations.}\label{results3}
\end{table}

The average precision, recall, and $F$ values in Table~\ref{results3} are calculated using 
macroaveraging, rather than microaveraging \cite{Lew}. Microaveraging combines 
the true positive, false positive, and false negative counts for all of the classes, and then 
calculates precision, recall, and $F$ from the combined counts. Macroaveraging calculates 
the precision, recall, and $F$ for each class separately, and then calculates the averages 
across all classes. Macroaveraging gives equal weight to all classes, but microaveraging 
gives more weight to larger classes. We use macroaveraging (giving equal weight to all 
classes), because we have no reason to believe that the class sizes in the data set reflect 
the actual distribution of the classes in a real corpus. (Microaveraging would give a slight 
boost to our results.)

We can adjust the balance between precision and recall, using a method similar to the 
approach in Section~\ref{subsubsec:recognizing}. For each noun-modifier pair that is to be classified, we find the 
two nearest neighbours. If the two nearest neighbours belong to the same class, then we 
output that class as our guess for the noun-modifier pair that is to be classified. 
Otherwise, we calculate the margin (the cosine of the first nearest neighbour minus the 
cosine of the second nearest neighbour). Let $m$ be the margin and let $t$ be the threshold. If 
$-m \le t \le +m$, then we output the class of the first nearest neighbour as our guess for the 
given noun-modifier pair. If $t > m$, then we abstain from classifying the given noun-modifier 
pair (we output no guess). If $t < -m$, then we output two guesses for the given 
noun-modifier pair, the classes of both the first and second nearest neighbours.

\begin{figure}[t]
\centerline{\includegraphics[width=\columnwidth,keepaspectratio]{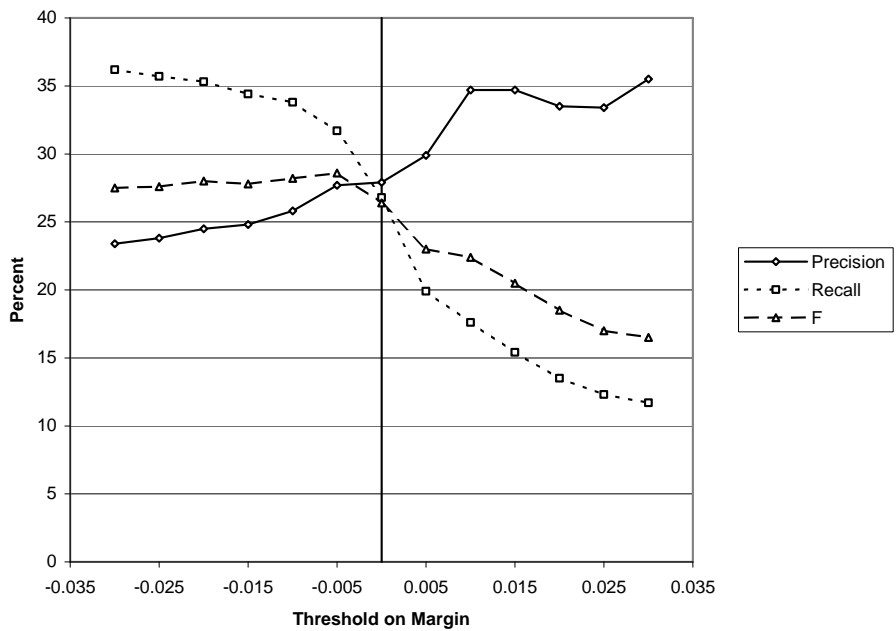}}
\caption{Precision, recall, and $F$ with varying thresholds on the margin, for 30 classes.}\label{figure2}
\end{figure}

Figure~\ref{figure2} shows the trade-off between precision and recall as the threshold on the margin 
varies from $-0.03$ to $+0.03$. The precision, recall, and $F$ values that are plotted here are 
the averages across the 30 classes. The vertical line at zero corresponds to the bottom row 
in Table~\ref{results3}. With a threshold of $+0.03$, precision rises to 35.5\% and recall falls to 11.7\%. 
With a threshold of $-0.03$, recall rises to 36.2\% and precision falls to 23.4\%. 

In Figure~\ref{figure2}, $F$ is higher for negative thresholds on the margin. We do not have an 
explanation for this. We believe it is due to noise. 

\subsubsection{Five Classes}\label{subsubsec:five}

Classification with 30 distinct classes is a hard problem. To make the task easier, we can 
collapse the 30 classes to 5 classes, using the grouping that is given in Table~\ref{classes}. For 
example, {\em agent} and {\em beneficiary} both collapse to {\em participant}. 
Table~\ref{results4} gives the results for 
the 5 class problem. Random guessing would yield an average $F$ value of 20.0\%, but the 
actual average $F$ value is 43.2\%. The difference (23.2\%) is significant with 95\% 
confidence ($p < 0.05$, according to the paired t-test). The accuracy is 45.7\% 
(274/600). 

\begin{table}[t]
\begin{tabular}{lrrrrr}
\hline
Class name       & Class size & Class percent & Precision & Recall   & $F$       \\
\hline
causality        & 86         & 14.3\%        & 21.2\%    & 24.4\%   & 22.7\%    \\
participant      & 260        & 43.3\%        & 55.3\%    & 51.9\%   & 53.6\%    \\
quality          & 146        & 24.3\%        & 45.4\%    & 47.3\%   & 46.3\%    \\
spatial          & 56         & 9.3\%         & 29.1\%    & 28.6\%   & 28.8\%    \\
temporality      & 52         & 8.7\%         & 66.0\%    & 63.5\%   & 64.7\%    \\
\hline
Average          & 120        & 20.0\%        & 43.4\%    & 43.1\%   & 43.2\%    \\
\hline
\end{tabular}
\vspace{-15pt}
\caption[]{The precision, recall, and $F$ for each of the 5 groups of classes of 
semantic relations.}\label{results4}
\end{table}

\begin{figure}[htb]
\centerline{\includegraphics[width=\columnwidth,keepaspectratio]{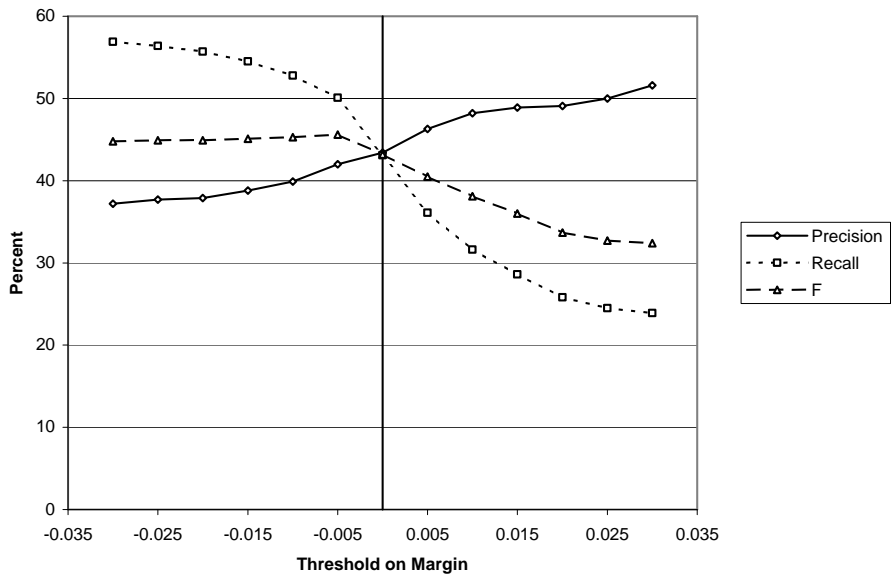}}
\caption{Precision, recall, and $F$ with varying thresholds on the margin, for 5 classes.}\label{figure3}
\end{figure}

As before, we can adjust the balance between precision and recall by varying a threshold 
on the margin. Figure~\ref{figure3} shows precision and recall as the threshold varies from $-0.03$ to 
$+0.03$. The precision, recall, and $F$ values are averages across the 5 classes (macroaverages). The vertical 
line at zero corresponds to the bottom row in Table~\ref{results4}. With a threshold of $+0.03$, 
precision rises to 51.6\% and recall falls to 23.9\%. With a threshold of $-0.03$, recall rises 
to 56.9\% and precision falls to 37.2\%.

These experiments required 76,800 queries to AltaVista (600 word pairs $\times$ 128 queries 
per word pair). With a five second delay between each query, processing the queries took 
about five days.

\subsection{Discussion}\label{subsec:nounmoddiscuss}

The performance of the nearest-neighbour VSM algorithm is well above random chance. 
With 30 classes, the average $F$ is 26.5\%, where random guessing would give an expected 
average $F$ of 3.3\%. With 5 classes, the average $F$ is 43.2\%, where random guessing would 
give an expected average $F$ of 20.0\%. As far as we know, this is the first attempt to 
classify semantic relations without a lexicon. Research with the same data \cite{NasSzp}, 
but using a lexicon, is still in the exploratory phase.

However, there is clearly much opportunity for improvement. Most practical tasks would 
likely require higher accuracy than we have obtained here. One place to look for 
improvement is in the joining terms. For the experiments in this section, we used the 
same joining terms as with the analogy questions (Table~\ref{joining}). It seems possible that the 
joining terms that work best for analogy questions are not necessarily the same as the 
terms that work best for classifying semantic relations. The kinds of semantic relations 
that are typically tested in SAT questions are not necessarily the kinds of semantic 
relations that typically appear in noun-modifier pairs. 

We also expect better results with more data. Although 600 noun-modifier pairs may 
seem like a lot, there are 30 classes, so the average class has only 20 examples. We would 
like to have at least 100 examples of each class, but manually labeling 3000 examples would 
require a substantial amount of painstaking effort. 

The classification scheme given in Table~\ref{classes} is only one of many possible
ways of classifying semantic relations. Each of the papers discussed in Section~\ref{subsec:semrels}
has a different classification of semantic relations (\citeauthor{Van}, \citeyear{Van};
\citeauthor{BarSzp}, \citeyear{BarSzp}; \citeauthor{RosHea}, \citeyear{RosHea}; 
\citeauthor{Ros}, \citeyear{Ros}; \citeauthor{NasSzp}, \citeyear{NasSzp}).
\citeauthor{Madetal} (\citeyear{Madetal}) give a carefully constructed
hierarchy of semantic relations, but this classification scheme has not yet
been applied to labeling noun-modifier pairs.
None of these classification schemes have been validated
by determining the level of inter-annotator agreement. 

Another limitation is the assumption that each noun-modifier pair can only
belong to one class. For example, ``concert hall'' might be classified
as {\em purpose} (Table~\ref{classes}), but it could equally well be
classified as {\em location}. A more flexible approach would allow multiple
labels for each noun-modifier pair.

It is reasonable to doubt that any classification scheme for semantic relations
can be complete. Each domain has its own special types of semantic relations.
For example, \citeauthor{Steetal} (\citeyear{Steetal}) provide a classification scheme for
relationships between genes, including classes such as ``$N\!P_0$ phosphorylates $N\!P_1$''.
However, it is plausible that a general-purpose scheme like Table~\ref{classes}
can capture the majority of semantic relations in general text at a reasonable
level of granularity.

\section{Limitations and Future Work}\label{sec:future}

Perhaps the biggest limitation of this work is the accuracy that we have achieved so far. 
Although it is state-of-the-art for SAT analogy questions and unrestricted-domain noun-modifier 
semantic relations, it is lower than we would like. However, both of these tasks 
are ambitious and research on them is relatively new. We believe that the results are 
promising and we expect significant improvements in the near future.

The VSM has been extensively explored in information retrieval. There are many ideas in 
the IR literature that might be used to enhance the performance of VSM applied to 
analogies and semantic relations. We have begun some preliminary exploration of 
various term weighting schemes \cite{SalBuc} and extensions of the VSM 
such as the GVSM \cite{Won} and LSA \cite{LanDum}. 

An area for future work is exploring the sensitivity of the VSM to the size of the
corpus. It seems plausible that our (limited) success with the VSM is due (to a 
large extent) to the huge corpus indexed by AltaVista. 
It is possible that the data we need, regarding relations between words, is highly 
sparse. Our approach might fail
with a typical corpus, such as the British National Corpus (BNC). We estimate
that AltaVista indexes about $10^{11}$ words, but BNC only contains about $10^{8}$ words.

However, more sophisticated algorithms, such as LSA, may be able to extract the
necessary information from a much smaller corpus. For the task of measuring similarity
between individual words, \citeauthor{LanDum} (\citeyear{LanDum}) compared the
cosine measure using vectors generated directly from a corpus versus vectors
generated by applying LSA to the corpus. On the TOEFL multiple-choice
synonym questions, the cosine measure with directly-generated vectors achieved a score of 
only 36.8\%, but the cosine measure with LSA-generated vectors achieved a score of 
64.4\%.\endnote{\citeauthor{LanDum} (\citeyear{LanDum}) report scores that were
corrected for guessing by subtracting a penalty of $1/3$ for each incorrect answer.
The performance of 64.4\% translates to 52.5\% when corrected for guessing, and 36.8\%
translates to 15.8\%.}

We believe that our set of joining terms (Table~\ref{joining}) is far from ideal. It seems 
likely that much larger vectors, with thousands of elements instead of 128, would improve the 
performance of the VSM algorithm. With the current state of technology, experiments 
with alternative sets of joining terms are very time consuming. 

The joining terms raise some interesting questions, which we have not yet addressed.
Which terms are most important? Many of them are prepositions. Does this work
have any significant implications for research in the semantics of prepositions \cite{Reg}?
Many of them are verbs. What are the implications for research in the semantics
of verbs \cite{GilJur}? Can we use any ideas from research on prepositions and verbs to guide
the search for an improved set of joining terms? These are questions for future work.

In this paper, we have focused on the VSM algorithm, but we believe that ensemble 
methods will ultimately prove to yield the highest accuracy \cite{Turetal}. 
Language is a complex, heterogeneous phenomenon, and it seems unlikely that any 
single, pure approach will be best. The best approach to analogies and semantic relations 
will likely combine statistical and lexical resources. However, as a research strategy, it 
seems wise to attempt to push the performance of each individual module as far as 
possible before combining the modules. 

\section{Conclusion}\label{sec:conclusion}

We believe that analogy and metaphor play a central role in human cognition and 
language \cite{LakJoh, Hof, Fre}. SAT-style 
analogy questions are a simple but powerful and objective tool for investigating these 
phenomena. Much of our everyday language is metaphorical, so progress in this area is 
important for computer processing of natural language. 

A more direct application of SAT question answering technology is classifying 
noun-modifier relations, which has potential applications in machine translation, information 
extraction, and word sense disambiguation. Contrariwise, a good algorithm for 
classifying semantic relations should also help to solve verbal analogies, which argues for 
a strong connection between recognizing analogies and classifying semantic relations. 

In this paper, we have shown how the cosine metric in the Vector Space Model can be 
used to solve analogy questions and to classify semantic relations. The VSM performs 
much better than random chance, but below human levels. However, the results indicate 
that these challenging tasks are tractable and we expect further improvements. We 
believe that the VSM can play a useful role in an ensemble of algorithms for learning 
analogies and semantic relations.

\acknowledgements

We are grateful to Vivi Nastase and Stan Szpakowicz for sharing their list of 600
classified noun-modifier phrases with us. Thanks to AltaVista for allowing us to send so
many queries to their search engine. Thanks to the anonymous referees of {\em Machine
Learning} for their helpful comments on an earlier version of this paper, and to the
editors, Pascale Fung and Dan Roth, for their work on preparing this special issue.

\theendnotes

\end{article}
\end{document}